\documentclass{article}

\usepackage[final]{corl_2021} 

\usepackage{amsmath,amsfonts,bm}

\def\eqref#1{equation~\ref{#1}}

\def\1{\bm{1}}

\def\rvs{{\mathbf{s}}}

\def\va{{\bm{a}}}

\def\vc{{\bm{c}}}
\def\vd{{\bm{d}}}

\def\vn{{\bm{n}}}
\def\vo{{\bm{o}}}

\def\vr{{\bm{r}}}

\def\vu{{\bm{u}}}
\def\vv{{\bm{v}}}

\def\vx{{\bm{x}}}

\DeclareMathAlphabet{\mathsfit}{\encodingdefault}{\sfdefault}{m}{sl}
\SetMathAlphabet{\mathsfit}{bold}{\encodingdefault}{\sfdefault}{bx}{n}

\usepackage{color}
\usepackage{epsfig}
\usepackage{graphicx}

\usepackage{adjustbox}
\usepackage{array}
\usepackage{booktabs}
\usepackage{colortbl}
\usepackage{float,wrapfig}
\usepackage{hhline}
\usepackage{multirow}

\usepackage{amsmath,amsfonts,amsthm,amssymb}
\usepackage{bm}
\usepackage{nicefrac}
\usepackage{microtype}

\usepackage{changepage}
\usepackage{extramarks}
\usepackage{fancyhdr}
\usepackage{lastpage}
\usepackage{setspace}
\usepackage{soul}
\usepackage{xspace}

\usepackage{url}

\usepackage{algorithm, algorithmic}
\usepackage{enumerate}
\usepackage{enumitem}
\usepackage{verbatim}

\newcolumntype{L}[1]{>{\raggedright\let\newline\\\arraybackslash\hspace{0pt}}m{#1}}
\newcolumntype{C}[1]{>{\centering\let\newline\\\arraybackslash\hspace{0pt}}m{#1}}
\newcolumntype{R}[1]{>{\raggedleft\let\newline\\\arraybackslash\hspace{0pt}}m{#1}}

\newcommand{\sect}[1]{Section~\ref{#1}}

\newcommand{\eqn}[1]{Equation~\ref{#1}}
\newcommand{\fig}[1]{Figure~\ref{#1}}

\newcommand{\ignore}[1]{}

\makeatletter
\DeclareRobustCommand\onedot{\futurelet\@let@token\@onedot}
\def\@onedot{\ifx\@let@token.\else.\null\fi\xspace}

\def\eg{e.g\onedot} 
\def\ie{i.e\onedot}

\makeatother

\definecolor{MyDarkBlue}{rgb}{0,0.08,1}
\definecolor{MyDarkGreen}{rgb}{0.02,0.6,0.02}
\definecolor{MyDarkRed}{rgb}{0.8,0.02,0.02}
\definecolor{MyDarkOrange}{rgb}{0.40,0.2,0.02}
\definecolor{MyPurple}{RGB}{111,0,255}
\definecolor{MyRed}{rgb}{1.0,0.0,0.0}
\definecolor{MyGold}{rgb}{0.75,0.6,0.12}
\definecolor{MyDarkgray}{rgb}{0.66, 0.66, 0.66}

\title{3D Neural Scene Representations for \\ Visuomotor Control}

\author{
  Yunzhu Li\thanks{equal contribution. Project Page: \url{https://3d-representation-learning.github.io/nerf-dy/}}\\
  MIT CSAIL\\
  \texttt{liyunzhu@mit.edu} \\
  \And
  Shuang Li\footnotemark[1]\\
  MIT CSAIL\\
  \texttt{lishuang@mit.edu} \\
  \And
  Vincent Sitzmann\\
  MIT CSAIL\\
  \texttt{sitzmann@mit.edu} \\
  \AND
  Pulkit Agrawal\\
  MIT CSAIL\\
  \texttt{pulkitag@mit.edu} \\
  \And
  Antonio Torralba\\
  MIT CSAIL\\
  \texttt{torralba@mit.edu}
}

\begin{document}
\maketitle

\begin{abstract}

Humans have a strong intuitive understanding of the 3D environment around us. The mental model of the physics in our brain applies to objects of different materials and enables us to perform a wide range of manipulation tasks that are far beyond the reach of current robots. In this work, we desire to learn models for dynamic 3D scenes purely from 2D visual observations. Our model combines Neural Radiance Fields (NeRF) and time contrastive learning with an autoencoding framework, which learns viewpoint-invariant 3D-aware scene representations. We show that a dynamics model, constructed over the learned representation space, enables visuomotor control for challenging manipulation tasks involving both rigid bodies and fluids, where the target is specified in a viewpoint different from what the robot operates on. When coupled with an auto-decoding framework, it can even support goal specification from camera viewpoints that are \emph{outside the training distribution}.
We further demonstrate the richness of the learned 3D dynamics model by performing future prediction and novel view synthesis. Finally, we provide detailed ablation studies regarding different system designs and qualitative analysis of the learned representations.

\end{abstract}
\section{Introduction}

Existing state-of-the-art model-based systems operating from vision typically treat images as 2D grids of pixels~\cite{ebert2018visual,hafner2019dream,schrittwieser2020mastering}. The world, however, is three-dimensional. Modeling the environment from 3D enables amodal completion and allows the agents to operate from different views.
Therefore, it is desirable to obtain good 3D-aware representations of the environment from 2D observations to achieve better task performance when an accurate inference of 3D information is essential, which can further make it easier to specify tasks, learn from third-person videos, etc.

One of the core questions of model learning in robotic manipulation is how to determine the state representation for learning the dynamics model.
The desired representation should make it easy to capture the environment dynamics, exhibit a good 3D understanding of the objects in the scene, and be applicable to diverse object sets such as rigid or deformable objects and fluids.
One line of prior work learns the dynamics model directly in the image pixel space~\cite{finn2016unsupervised,ebert2017self,yen2020experience,suh2020surprising}. However, modeling dynamics in such a high-dimensional space is challenging, and these methods typically generate blurry images when performing the long-horizon future predictions.
Another line of work focused on only predicting task-relevant features identified as keypoints~\cite{manuelli2019kpam,manuelli2020keypoints,wang20206,gao2021kpam,li2020causal}. Such models perform well in terms of category-level generalization, \ie, the same set of keypoints can represent different instances within the same category, but are not sufficient to model objects with large variations like fluids and granular materials.
Other methods learn dynamics in the latent space~\cite{watter2015embed,agrawal2016learning,hafner2019learning,hafner2019dream,schrittwieser2020mastering}. 
However, the majority of these methods learn dynamics models using 2D convolutional neural networks and reconstruction loss -- which has the same problem as predicting dynamics in the image space, \ie, their learned representations lack equivariance to 3D transformations.
Time contrastive networks~\cite{sermanet2018time}, on the other hand, aim to learn viewpoint-invariant representations from multi-view inputs, but do not require detailed modeling of 3D contents.
As a result, previously unseen scene configurations and camera poses are out-of-distribution samples for the state estimator. As we will see, this leads to wrong state estimates and results in faulty control trajectories. 

\begin{figure}[t]
    \begin{center}
        \includegraphics[width=\linewidth]{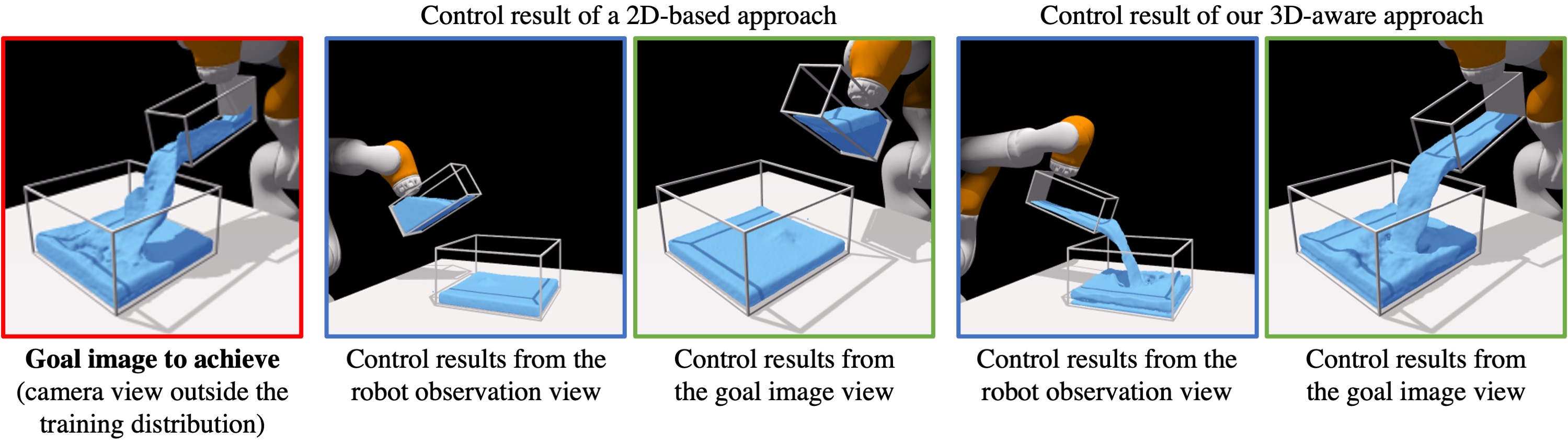}
    \end{center}
    \vspace{-8pt}
    \caption{\small {\bf Comparison of the control results between a 2D-based baseline and our 3D-aware approach.}
    The task here is to achieve the configuration shown on the left, observed from a viewpoint that is outside the training distribution.
    The agent only takes a single-view visual observation as input (images with blue frames) from a viewpoint that is vastly different from the goal. Our method generalizes well in this scenario and outperforms the 2D-based baseline, demonstrating the benefits of the learned 3D-aware scene representations.
    }
    \vspace{-5pt}
    \label{fig:teaser}
\end{figure}

Meanwhile, recent work in computer vision has made impressive progress in the learning of 3D-structured neural scene representations. These approaches allow inference of 3D structure and appearance, trained only given 2D observations, either by overfitting on a single scene~\cite{sitzmann2018deepvoxels,lombardi2019neural,mildenhall2020nerf} or by generalizing across scenes~\cite{sitzmann2019scene,tung2019learning,tung20203d}.
Through their 3D inductive bias, the scene representations inferred by these models encode the scene contents with better accuracy and are invariant to changes in camera perspectives.
It is desirable to push these ideas further to obtain a deeper understanding of how these methods, which directly reason over 3D, can bring in new characteristics and how they can be beneficial for dynamics modeling and complicated control tasks.

In this work, we aim to leverage recently proposed 3D-structure-aware implicit neural scene representations for visuomotor control tasks. We thus propose to embed neural radiance fields~\cite{mildenhall2020nerf} in an auto-encoder framework, enabling tractable inference of the 3D-structure-aware scene state for dynamic environments.
By additionally enforcing a time contrastive loss on the estimated states, we ensure that the learned state representations are viewpoint-invariant.
We then train a dynamics model that predicts the evolution of the state space conditioned on the input action, enabling us to perform control in the learned state space.
Though the representation itself is grounded in the 3D implicit field, the convolutional encoder is not.
At test time, we overcome this limitation by performing inference-via-optimization~\cite{park2019deepsdf,sitzmann2019scene}, enabling accurate state estimation even for out-of-distribution camera poses and, therefore, control of tasks where the goal view is specified in an entirely unseen camera perspective. These contributions enable us to perform model-based visuomotor control of complex scenes, modeling 3D dynamics of both rigid objects and fluids.
Through comparison with various baselines, the learned representation from our model is more precise at describing the contents of 3D scenes, which allows it to accomplish control tasks involving rigid objects and fluids with significantly better accuracy (\fig{fig:teaser}).
Please see our supplementary video for better visualization.

We summarize our contributions as follows:
(i)~We extend an autoencoding framework with a neural radiance field rendering module and time contrastive learning that allows us to learn 3D-aware scene representations for dynamics modeling and control purely from visual observations.
(ii)~By incorporating the auto-decoder mechanism at test time, our framework can adjust the learned representation and accomplish the control tasks with the goal specified from camera viewpoints outside the training distribution.
(iii)~We are the first to augment neural radiance fields using a time-invariant dynamics model, supporting future prediction and novel view synthesis across a wide range of environments with different types of objects.

\begin{figure*}
    \begin{center}
        \includegraphics[width=\linewidth]{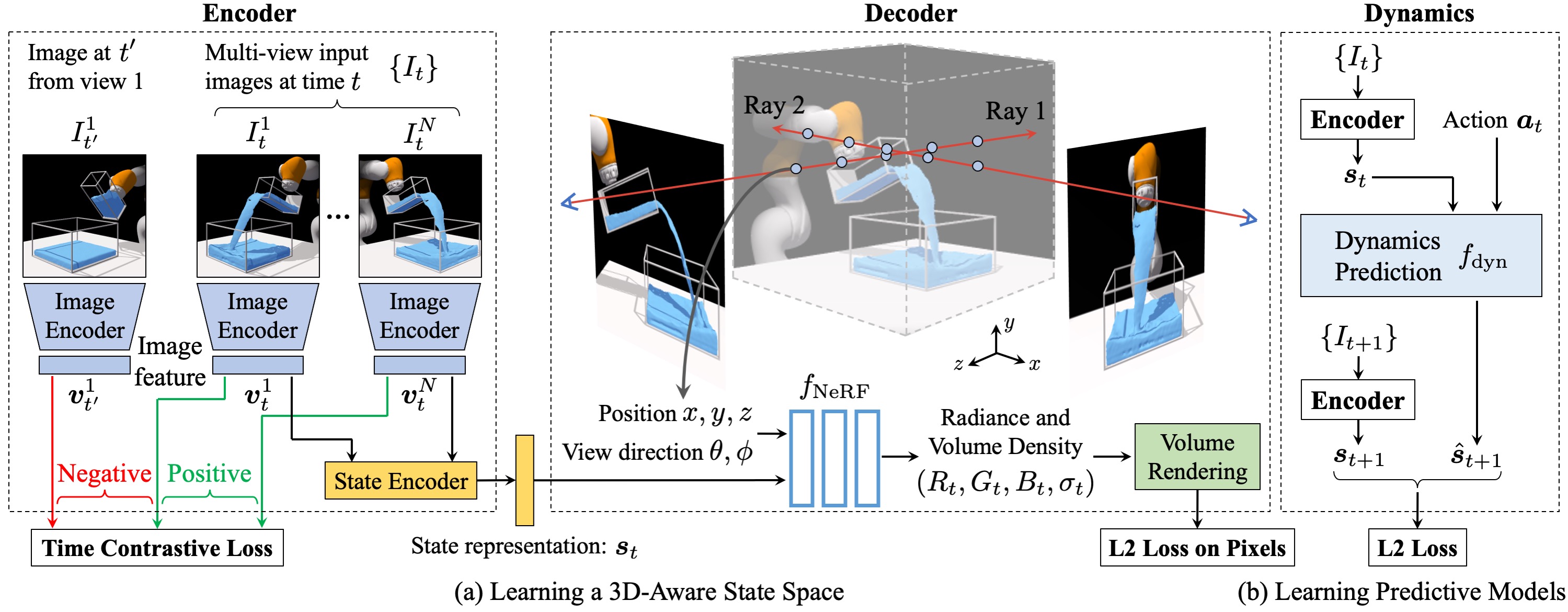}
    \end{center}
    \vspace{-8pt}
    \caption{\small{{\bf Overview of the training procedure.} 
    {\bf Left:} an encoder that maps the input images into a latent scene representation. The images are first sent to an image encoder to generate the image feature representations $\vv$. Then we combine the image features from the same time step using a state encoder to obtain the state representation $\bm{s}_t$. 
    A time contrastive loss is applied to enable our model to be invariant to camera viewpoints. 
    {\bf Middle:} a decoder that takes the scene representation as input and generates the visual observation conditioned on a given viewpoint. We use an L2 loss to ensure the reconstructed image to be similar to the ground truth image.
    {\bf Right:} a dynamics model that predicts the future scene representations $\hat{\bm{s}}_{t+1}$ by taking in the current representation $\bm{s}_t$ and action $\bm{a}_t$. We use an L2 loss to enforce the predicted latent representation to be similar to the scene representation $\bm{s}_{t+1}$ extracted from the true visual observation $I_{t+1}$.
    }
    }
    \vspace{-5pt}
    \label{fig:overview_train}
\end{figure*}

\vspace{-2pt}
\section{Related Work}
\vspace{-3pt}
\textbf{3D Scene Representation Learning.}
Prior work leverages the latent spaces of autoencoder-like models as learned representations of the underlying 3D scene to enable novel view synthesis from a single image~\citep{worrall2017interpretable,tatarchenko2015single}.
\citet{eslami2018neural} embed this approach in a probabilistic framework.
To endow models with 3D structure, voxelgrids can be leveraged as neural scene representations~\citep{rezende2016unsupervised,nguyen2018rendernet,sitzmann2018deepvoxels,tung2018learning,zhu2018visual}, while others have tried to predict particle sets from images~\cite{li2020visual} or embed an explicit 3D representation to enable inference from never-before-seen viewpoints~\cite{saponaro2019beyond}.
\citet{sitzmann2019scene} propose to learn neural implicit representations of 3D shape and appearance supervised with posed 2D images via a differentiable renderer.
Generalizing across neural implicit representations can also be realized by local conditioning on CNN features~\cite{saito2019pifu,trevithick2020grf,yu2020pixelnerf}, but this does not learn a global representation of the scene state.
Alternatively, gradient-based meta-learning has been proposed for faster inference of implicit neural representations~\cite{sitzmann2020metasdf}.
Deformable scenes can be modeled by transporting input coordinates to neural implicit representations with an implicitly represented flow field or time-variant latent code~\cite{niemeyer2019occupancy,park2020deformable,pumarola2020d,tretschk2020non,li2020neural,du2020neural,xian2021space,ost2021neural,li2021neural}; however, they typically fit one trajectory and cannot handle different initial conditions and external action inputs, limiting their use in control.

\textbf{Model-Based RL in Robotic Manipulation.}
We can categorize model-based RL methods by whether they use physics-based or data-driven models, and whether they assume full state access or only visual observation.
Methods that rely on physics-based models typically assume full-state information of the environment~\cite{hogan2016feedback,zhou2019pushing} and require the knowledge of the object models, making them hard to generalize to novel objects or partially observable scenarios.
For data-driven models, people have attempted to learn a dynamics model for closed-loop planar pushing~\citep{bauza2018data} or dexterous manipulation~\citep{nagabandi2020deep}.
\citet{schenck2017reasoning,schenck2018perceiving} tackle a similar fluid pouring task via closed-loop simulation.
Although they have achieved impressive results, they rely on state estimators customized for specific tasks, limiting their applicability to more general and diversified manipulation tasks.

Various model-based RL methods have been proposed to learn state representations from visual observations, such as image-space dynamics~\cite{finn2016unsupervised,ebert2017self,ebert2018visual,suh2020surprising}, keypoint representation~\cite{kulkarni2019unsupervised,manuelli2020keypoints,li2020causal}, and low-dimensional latent space~\cite{watter2015embed,hafner2019learning,hafner2019dream,schrittwieser2020mastering}.
Some works learn a meaningful representation space using reconstruction loss~\cite{hafner2019learning,hafner2019dream}. Others jointly train the forward and inverse dynamics models~\cite{agrawal2016learning}, or use time contrastive loss to regularize the latent embedding~\cite{sermanet2018time}.
We differ from the previous methods by explicitly incorporating a 3D volumetric rendering process during training.

\vspace{-2pt}
\section{3D-Aware Representation Learning for Dynamics Modeling}
\vspace{-3pt}
Inspired by Neural Radiance Fields (NeRF)~\cite{mildenhall2020nerf}, we propose a framework that learns a viewpoint-invariant model for dynamic environments. 
As shown in \fig{fig:overview_train}, our framework has three parts: (1)~an encoder that maps the input images into a latent state representation, 
(2)~a decoder that generates an observation image under a certain viewpoint based on the state representation,
and (3)~a dynamics model that predicts the future state representations based on the current state and the input action.

\vspace{-2pt}
\subsection{3D-Aware Scene Representation Learning}
\vspace{-3pt}

\textbf{Neural Radiance Field.}
\label{sec:nerf}
Given a 3D point $\vx \in \mathbb{R}^3$ in a scene and a viewing direction unit vector $\vd \in \mathbb{R}^3$ from a camera, NeRF learns to predict a volumetric radiance field. This is represented using a differentiable rendering function $f_\text{NeRF}$ that predicts the corresponding density $\sigma$ and RGB color $\vc$ using $f_\text{NeRF}(\vx, \vd) = (\sigma, \vc)$.
To render the color of an image pixel, NeRF integrates the information along the camera ray using
$\hat{\boldsymbol{C}}(\vr) = \int_{h_\text{near}}^{h_\text{far}}{T(h)\sigma(h)\vc(h)}dh$, 
where $\vr(h)=\vo + h\vd$ is the camera ray with its origin $\vo \in \mathbb{R}^3$ and unit direction vector $\vd \in \mathbb{R}^3$, and
$T(h)=\text{exp}(-\int_{h_\text{near}}^{h}{\sigma(s)}ds)$ is the accumulated transparency between the pre-defined near depth $h_\text{near}$ and far depth $h_\text{far}$ along that camera ray.
The mean squared error between the reconstructed color $\hat{\boldsymbol{C}}$ and the ground truth $\boldsymbol{C}$ is:
\begin{equation}
\small
    \mathcal{L}_{\text{rec}} = \sum_{\vr}{\| \hat{\boldsymbol{C}}(\vr) - {\boldsymbol{C}}(\vr)\|_2^2}.
    \label{eqn:nerf_loss}
\end{equation}
\textbf{Neural Radiance Field for Dynamic Scenes.}
\label{sec:nerf_dynamic}
One key limitation of NeRF is that it assumes the scene is static. For a dynamic scene, it must learn a separate radiance field $f_\text{NeRF}$ for each time step.
This severely limits the ability of NeRF to be used in planning and control, as it is unable to handle dynamic scenes with different initial configurations or input action sequences.
While other models have shown generalization across scenes~\cite{sitzmann2019scene,niemeyer2020differentiable}, it's unclear how they can be used in visuomotor control.
To enable $f_\text{NeRF}$ to model dynamic scenes, we learn an encoding function $f_\text{enc}$ that maps the visual observations to a feature representation $\bm{s}$ for each time step and learn the volumetric radiance field decoding function based on $\bm{s}$.
Let $\{I_t\}$ denotes the set of 2D images that capture a 3D scene at time $t$ from one or more camera viewpoints. The image taken from the $i^\text{th}$ viewpoint is represented as $I^i_t$. 
We use ResNet-18~\cite{He_2016_CVPR} to extract a feature vector for each image. 
We take the output of ResNet-18 before the pooling layer and send it to a fully-connected layer, resulting in a 256 dimension image feature $\vv_t^i$.
This image feature is concatenated with the corresponding camera viewpoint information (a 16-D vector obtained by flattening the camera view matrix) and processed using a small multilayer perceptron (MLP) to generate the final image feature.
The scene representation $\bm{s}_t$ at time $t$ is generated by first averaging the image features across multiple camera viewpoints, then being encoded using another small MLP and normalized to have a unit L2 norm.

Given a 3D point $\vx$, a viewing direction unit vector $\vd$, and a scene representation $\bm{s}_t$, we learn a function $f_\text{dec}(\vx, \vd, \bm{s}_t)=(\sigma_t, \vc_t)$ to predict the radiance field represented by the density $\sigma_t$ and RGB color $\vc_t$. 
Similar to NeRF, we use the integrated information along the camera ray to render the color of image pixels from an input viewpoint and then compute the image reconstruction loss using \eqn{eqn:nerf_loss}.
During each training iteration, we render two images from different viewpoints to calculate more accurate gradient updates.
$f_\text{dec}$ depends on the scene representation $\bm{s}_t$, forcing it to encode the 3D contents of the scene to support rendering from different camera poses.

\begin{figure*}
    \begin{center}
        \includegraphics[width=\linewidth]{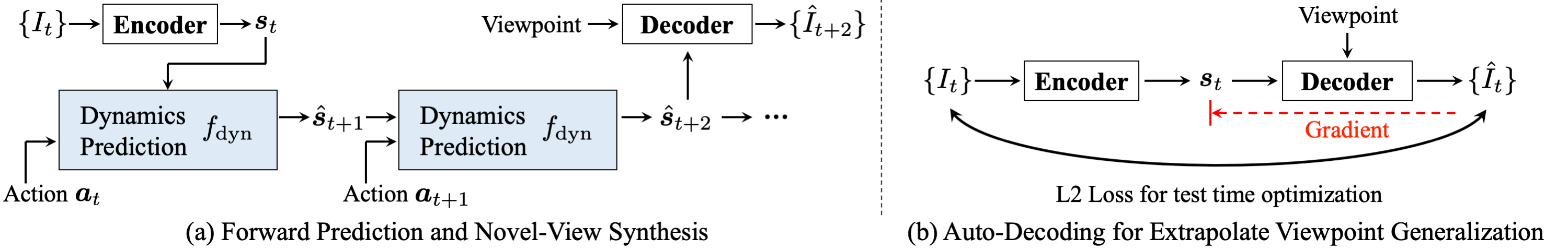}
    \end{center}
    \vspace{-8pt}
    \caption{\small {\bf Forward prediction and viewpoint extrapolation.}
    (a) We first feed the input image(s) at time $t$ to the encoder to derive the scene representation $\bm{s}_t$. 
    The dynamics model then takes $\bm{s}_t$ and the corresponding action sequence as input to iteratively predict the future.  The decoder synthesizes the visual observation conditioned on the predicted state representation and an input viewpoint.
    (b) We propose an auto-decoding inference-via-optimization framework to enable extrapolated viewpoint generalization. 
    Given an input image $I_t$ taken from a viewpoint outside the training distribution, 
    the encoder first predicts the scene representation $\bm{s}_t$. 
    Then the decoder reconstructs the observation $\hat{I}_t$ from $\bm{s}_t$ and the camera viewpoint from $I_t$. We calculate the L2 distance between $I_t$ and $\hat{I}_t$ and backpropagate the gradient through the decoder to update $\bm{s}_t$. The updating process is repeated for $K$ iterations, resulting in a more accurate $\bm{s}_t$  of the underlying 3D scene.
    }
    \vspace{-5pt}
    \label{fig:overview_test}
\end{figure*}

\textbf{Time Contrastive Learning.}
\label{sec:tc}
To enable the image encoder to be viewpoint invariant, we regularize the feature representation of each image $\vv^i_t$ using multi-view time contrastive loss (TCN) \cite{sermanet2018time} (see \fig{fig:overview_train}a).
The TCN loss encourages features of images from different viewpoints at the same time step to be similar, while repulsing features of images from different time steps to be dissimilar. 
More specifically, given a time step $t$, we randomly select one image $I_t^i$ as the anchor and extract its image feature $\vv_t^i$ using the image encoder.
Then we randomly select one positive image from the same time step but different camera viewpoint $I_t^{i'}$ and one negative image from a different time step but the same viewpoint $I_{t'}^i$. We use the same image encoder to extract their image features $\vv_t^{i'}$ and $\vv_{t'}^i$.
Similar to \cite{sermanet2018time}, we minimize the following time contrastive loss:
\begin{equation}
\small
    \mathcal{L}_\text{tc} = \max{(\| \vv_t^i - \vv_t^{i'} \|_2^2 - \| \vv_t^i - \vv_{t'}^i \|_2^2 + \alpha, 0)},
    \label{eqn:tc_loss}
\end{equation}
where $\alpha$ is a hyper-parameter denoting the \textit{margin} between the positive and negative pairs.

\vspace{-2pt}
\subsection{Learning the Predictive Model}
\vspace{-3pt}
After we have obtained the latent state representation $\bm{s}$, we use supervised learning to estimate the forward dynamics model, $\hat{\bm{s}}_{t+1} = f_\text{dyn}(\bm{s}_t, \bm{a}_t)$. 
Given $\bm{s}_t$ and
a sequence of actions $\{\va_t, \va_{t+1}, \dots \}$, we predict $H$ steps in the future by iteratively feeding in actions into the one-step forward model. 
We implement $f_\text{dyn}$ as an MLP network which is trained by optimizing the following loss function:  
\begin{equation}
\small
    \mathcal{L}_\text{dyn} = \sum_{h=1}^H \| \hat{\bm{s}}_{t+h} - \bm{s}_{t+h} \|_2^2, \;\;\; \text{where} \;\;\; 
    \hat{\bm{s}}_{t+h} = f_\text{dyn}(\hat{\bm{s}}_{t+h-1}, \bm{a}_{t+h-1}), 
    \;\;\; \hat{\bm{s}}_t = \bm{s}_t.
\end{equation}
We define the final loss as a combination of the image reconstruction loss, the time contrastive loss, and the dynamics prediction loss:
$\mathcal{L} = \mathcal{L}_\text{rec} + \mathcal{L}_\text{tc} + \mathcal{L}_\text{dyn}$.
We first train the encoder $f_\text{enc}$ and decoder $f_\text{dec}$ together using stochastic gradient descent (SGD) by minimizing $\mathcal{L}_\text{rec}$ and $\mathcal{L}_\text{tc}$, which makes sure that the learned scene representation $\bm{s}$ encodes the 3D contents and is viewpoint-invariant.
We then fix the encoder parameters, and train the dynamics model $f_\text{dyn}$ by minimizing $\mathcal{L}_\text{dyn}$ using SGD.
Please see the supplementary materials for the network architecture and training details.

\vspace{-2pt}
\section{Visuomotor Control}
\vspace{-3pt}
\subsection{Online Planning for Closed-Loop Control}
\label{sec:control}
\vspace{-3pt}
When given the goal image $I^\text{goal}$ and its associated camera pose, we can feed them through the encoder $f_\text{enc}$ to get the state representation for the goal configuration $\bm{s}^\text{goal}$. 
We use the same method to compute the state representation for the current scene configuration $\bm{s}_1$. The goal of the online planning problem is to find an action sequence $\va_1, \dots, \va_{T-1}$ that minimizes the distance between the predicted future representation and the goal representation at time $T$.
As shown in \fig{fig:overview_test}a, given a sequence of actions, our model can iteratively predict a sequence of latent state representations.
The latent-space dynamics model can then be used for downstream closed-loop control tasks via online planning with model-predictive control (MPC).
We formally define the online planning problem as follows:
\begin{equation}
\begin{aligned}
\small
    \min_{\va_{1}, \dots, \va_{T-1}} \quad & \| \bm{s}^\text{goal} - \hat{\bm{s}}_T \|_2^2, 
    \quad &
    \textrm{s.t.} \quad & \hat{\bm{s}}_1 = \bm{s}_1, \hat{\bm{s}}_{t+1} = f_\text{dyn}(\hat{\bm{s}}_t, \bm{a}_t).
\end{aligned}
\label{eqn:planning}
\end{equation}
Many existing off-the-shelf model-based RL methods can be used to solve the MPC problem~\cite{ebert2017self,nagabandi2020deep,hafner2019learning,finn2016unsupervised,manuelli2020keypoints,li2019learning,li2019propagation}.
We experimented with random shooting, gradient-based trajectory optimization, cross-entropy method, and model-predictive path integral (MPPI) planners~\cite{williams2015model} and found that MPPI performed the best.
In our experiments, we specify the action space as the position and orientation of the arm's end-effector. Then, the joint angle of the arm is calculated via inverse kinematics.
Please check our supplementary materials for more details on how we solve the planning task.

\begin{figure*}
    \begin{center}
        \includegraphics[width=\linewidth]{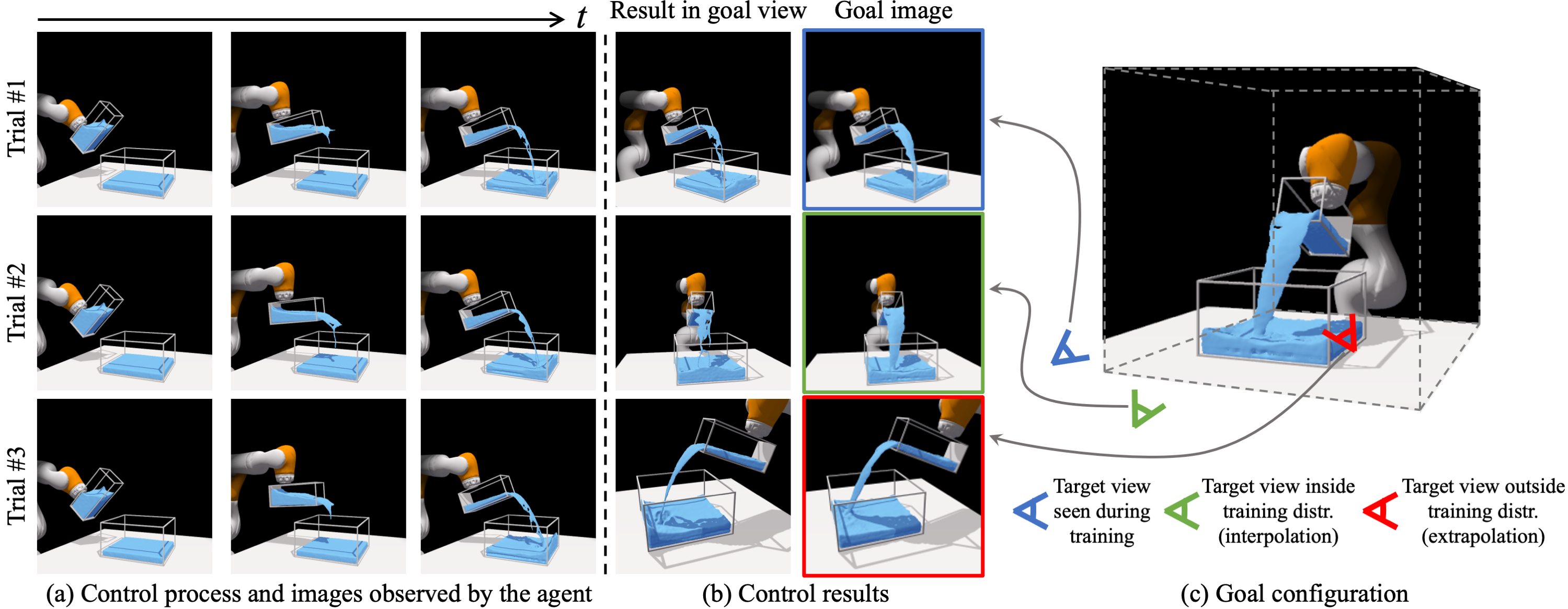}
    \end{center}
    \vspace{-8pt}
    \caption{\small{ {\bf Qualitative control results of our method on three types of testing scenarios.}
    The image on the right shows the target configuration we aim to achieve. The left three columns show the control processes, which are also the input images to the agent. The fourth column is the control results from the same viewpoint as the goal image. Trial $\#1$ specifies the goal using a different viewpoint from the agent's but has been encountered during training. Trial $\#2$ uses a goal view that is an interpolation of training viewpoints. Trial $\#3$ uses an extrapolated viewpoint that is outside the training distribution. Our method performs well in all settings.
    }
    }
    \label{fig:quali_control_process}
\end{figure*}
\begin{figure*}
    \begin{center}
        \includegraphics[width=\linewidth]{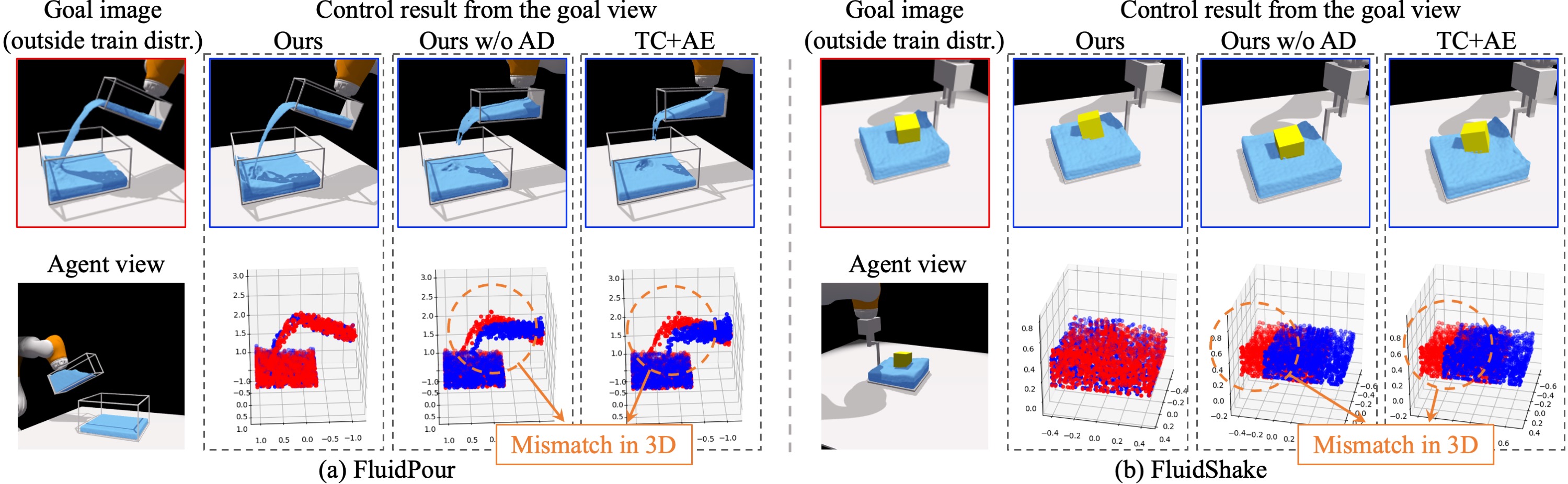}
    \end{center}
    \vspace{-8pt}
    \caption{\small{ {\bf Qualitative comparisons between our method and baseline approaches on the control tasks.}
    We show the closed-loop control results on the FluidPour and FluidShake environments. The goal image viewpoint (top-left image of each block) is outside the training distribution and is different from the viewpoint observed by the agent (bottom-left image of each block).
    Our final control results are much better than a variant that does not perform auto-decoding test-time optimization (Ours w/o AD) and the best-performing baseline (TC+AE), both of which fail to accomplish the task and their control results (blue points) exhibit an apparent deviation from the target configuration (red points) when measured in the 3D points space of the fluids and floating cube.
    }
    }
    \vspace{-5pt}
    \label{fig:quali_control_compare}
\end{figure*}

\vspace{-2pt}
\subsection{Auto-Decoder for Viewpoint Extrapolation}
\label{sec:autodecoder}
\vspace{-3pt}
End-to-end visuomotor agents can undergo significant performance drop when the test-time visual observations are captured from camera poses outside the training distribution.
The convolutional image encoder suffers from the same problem as it is not equivariant to changes in the camera pose, meaning it has a hard time generalizing to out-of-distribution camera views.
As shown in \fig{fig:overview_test}b, when encountered an image from a viewpoint outside training distribution, with a pixel distribution vastly different from what the model is trained on, passing it through the encoder $f_\text{enc}$ will give us an amortized estimation of the scene representation $\bm{s}_t$. 
It has a high chance that the decoded image is different from the ground truth as the viewpoint has never been encountered during training.

We fix this problem at test time by applying the inference-by-optimization (also named as an auto-decoding) framework that backpropagates through the volumetric renderer and the neural implicit representation into the state estimate~\cite{park2019deepsdf,sitzmann2019scene}.
This is inspired by the fact that the rendering function, $f_\text{dec}(\vx, \vd, \bm{s}_t) = (\sigma_t, \vc_t)$ is viewpoint equivariant, where the output only depends on the state representation $\bm{s}_t$, the location $\vx$, and the ray direction $\vd$, meaning that the output is invariant to the camera position along the camera ray, \ie, even we move the camera closer or farther away along the camera ray, $f_\text{dec}$ still tends to generate the same results.
We leverage this property and calculate the L2 distance between the input image and the reconstructed image $ \mathcal{L}_\text{ad} = \| I_t - \hat{I}_t \|^2_2$, and then update the scene representation $\bm{s}_t$ using stochastic gradient descent. We repeat this updating process $K$ times to derive the state representation of the underlying 3D scene. Note that this update only changes the scene representation while keeping the parameters in the decoder fixed.
The resulting representation is used as $\bm{s}^\text{goal}$ in \eqn{eqn:planning} to solve the online planning problem.

\begin{figure*}[t]
    \begin{center}
        \includegraphics[width=\linewidth]{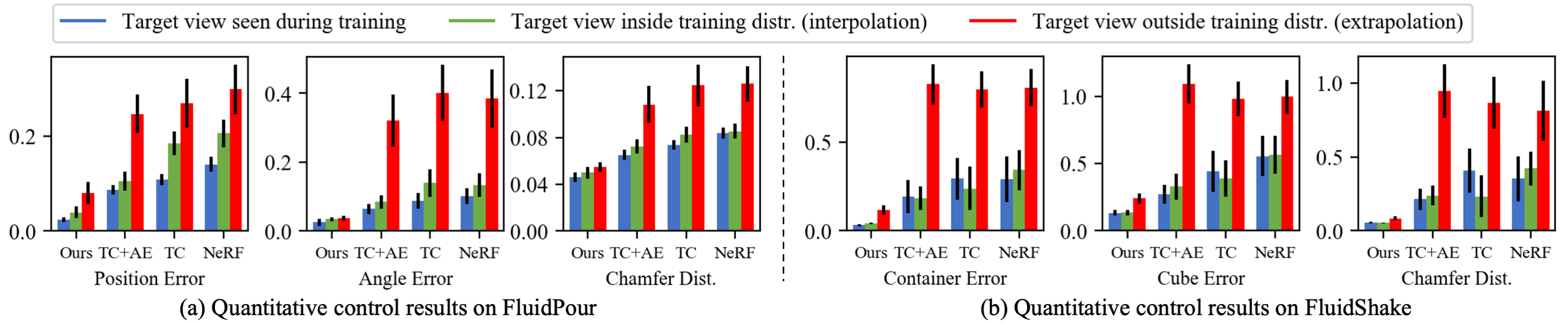}
    \end{center}
    \vspace{-8pt}
    \caption{\small {\bf Quantitative comparisons between our method and baselines on the FluidPour and FluidShake.}
    In each environment, we compare the results using three different evaluation metrics under three settings, \ie, (1) the target image view seen during training, (2) the target image view is inside the training distribution but not seen during training (interpolation), and (3) the target image view is outside the training distribution (extrapolation).
    The height of the bars indicates the mean, and the error bar denotes the standard error. Our model significantly outperforms all baselines under all testing settings.
    }
    \vspace{-5pt}
    \label{fig:quanti_control}
\end{figure*}

\vspace{-2pt}
\section{Experiments}
\vspace{-3pt}
\label{sec:exp}

\begin{figure*}[t]
    \begin{center}
        \includegraphics[width=\linewidth]{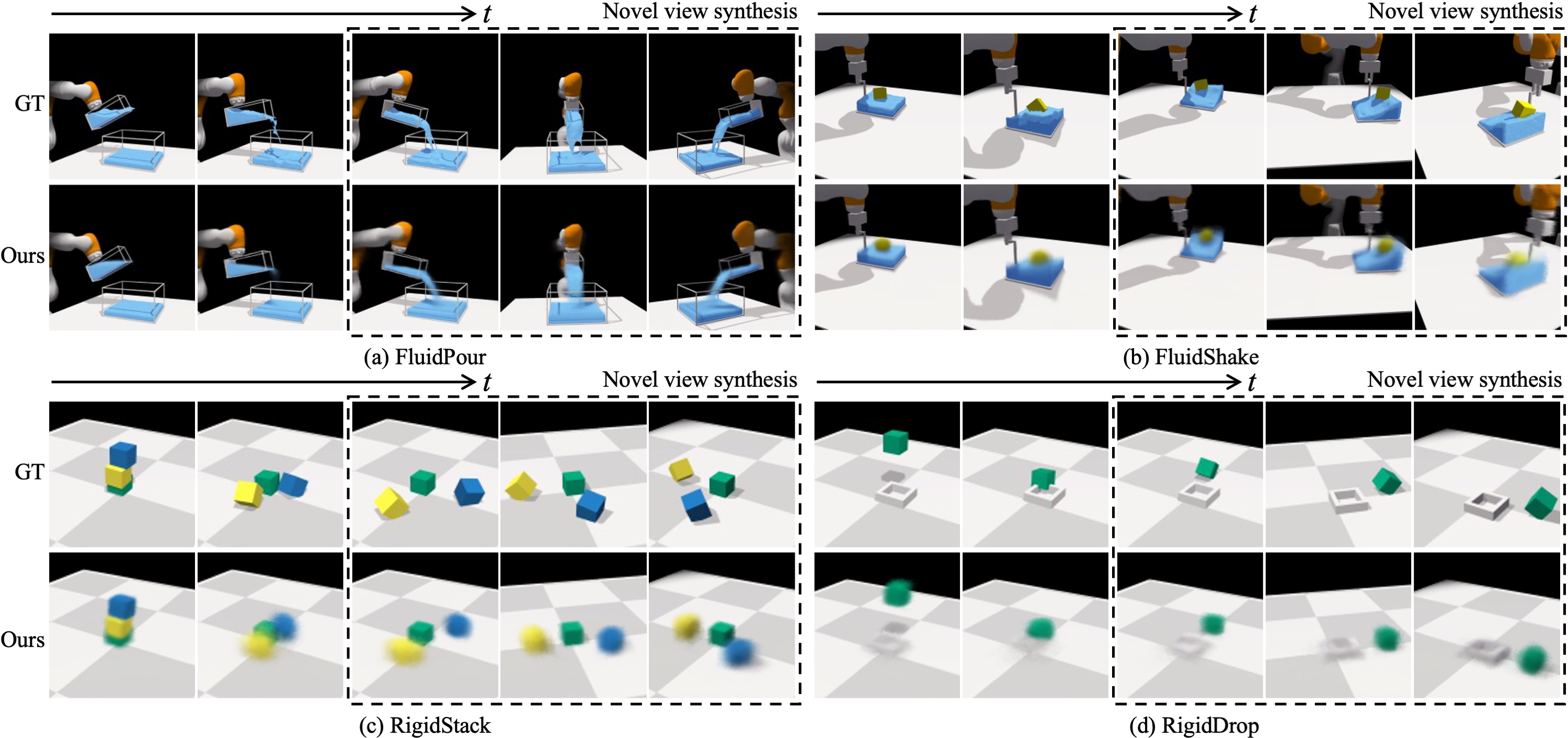}
    \end{center}
    \vspace{-9pt}
    \caption{\small{{\bf  Forward prediction and novel view synthesis on four environments.}
    Given a scene representation and an input action sequence, our dynamics model predicts the subsequent latent scene representation, which is used as the input of our decoder model to reconstruct the corresponding visual observation based on different viewpoints.
    In each block, we render images based on the open-loop future dynamic prediction from left to right.
    Images in the dotted box compare our model's novel view synthesis results in the last time step with the ground truth from three different viewpoints.
    }
    }
    \vspace{-9pt}
    \label{fig:quali_dynamics}
\end{figure*}

\textbf{Environments.}
We consider the following four environments involving both fluid and rigid objects to evaluate the proposed model and baseline approaches.
The environments are simulated using NVIDIA FleX~\cite{macklin2014unified}.
(1) FluidPour (\fig{fig:quali_dynamics}a): This environment contains a fully-actuated cup that pours fluids into a container at the bottom.
(2) FluidShake (\fig{fig:quali_dynamics}b): A fully-actuated container moves on a 2D plane. Inside the container are fluids and a rigid cube floating on the surface.
(3) RigidStack (\fig{fig:quali_dynamics}c): Three rigid cubes form a vertical stack and are released from a certain height but in different horizontal positions. They fall down and collide with each other and the ground.
(4) RigidDrop (\fig{fig:quali_dynamics}d): A cube falls down from a certain height. There is a container fixed at a random position on the ground. The cube either falls into the container or bounces out.
We use $20$ camera views for all environments and generated $1,000$ trajectories of $300$ time steps for both FluidPour and FluidShake as an offline dataset to train the model, $800$ trajectories of $80$ time steps for RigidStack, and $1,000$ trajectories of $50$ time steps for RigidDrop.

\textbf{Evaluation Metrics.}
We use the first two environments, \ie, FluidPour and FluidShake, to measure the control performance, where we specify the target configuration of the control task using images from (1) one of the viewpoints encountered during training, (2) an interpolated viewpoint between training viewpoints, and (3) an extrapolated viewpoint outside the training distribution (\fig{fig:quali_control_process}).

We provide quantitative evaluations on the control performance in FluidPour and FluidShake by extracting the particle set from the simulator, and measuring the Chamfer distance between the result and the goal, which we denote as ``Chamfer Dist.".
In FluidPour, we provide additional measurements on the L2 distance of the position/orientation of the cup towards the goal, denoting as ``Position Error" and ``Angle Error" respectively. In FluidShake, we calculate the L2 distance of the container and cube's position towards the goal and denote them as ``Container Error" and ``Cube Error" respectively.

\vspace{-2pt}
\subsection{Baseline Methods}
\vspace{-3pt}
\label{exp:baselines}
For comparison, we consider the following three baselines:
{\bf TC}: Similar to TCN~\cite{sermanet2018time}, it only uses time contrastive loss for learning the image feature without having to reconstruct the scene. We learn a dynamics model directly on the image features for control.
{\bf TC+AE}: Instead of using Neural Radiance Fields to reconstruct the image, this method uses the standard convolutional decoder to reconstruct the target image when given a new viewpoint. This would then be similar to GQN~\cite{eslami2018neural} augmented with a time contrastive loss.
{\bf NeRF}: This method is a direct adaptation from the original NeRF paper~\cite{mildenhall2020nerf} and is the same as ours except that it does not include the time contrastive loss during training and the auto-decoding test-time optimization.
We use the same dynamics model shown in \fig{fig:overview_train}b and train the model for each baseline respectively for dynamic prediction. 
We use the same feedback control method, \ie, MPPI \cite{williams2015model}, for our model and all the baselines.

\vspace{-2pt}
\subsection{Control Results}
\vspace{-3pt}
\label{exp:control_results}
\textbf{Goal Specification from Novel Viewpoints.}
\label{exp:control_nove_view}
Figure~\ref{fig:quali_control_process}c shows the goal configuration, and we ask the learned model to perform three control trials where the goal is specified from different types of viewpoints. 
The left three columns show the MPC control process from the agent's viewpoint. The fourth column visualizes the final configuration the agent achieves from the same viewpoint as the goal image. Trial $\#1$ specifies the goal using a different viewpoint from the agent's but has been encountered during training. Trial $\#2$ uses a goal view that is an interpolation of training viewpoints. Our agent can achieve the target configuration with decent accuracy.
For trial $\#3$, we specify the goal view by moving the camera closer, higher, and more downwards with respect to the container. Note this goal image view is outside the distribution of training viewpoints. With the help of test-time auto-decoding optimization introduced in \sect{sec:autodecoder},
our method can successfully achieve the target configuration, as shown in the figure.

\textbf{Baseline Comparisons.}
\label{exp:compare_baseline}
We benchmark our model with the baselines by assessing their performance on the downstream control tasks.
\fig{fig:quali_control_compare} shows the qualitative comparison between our model (Ours), a variant of our model that does not perform the auto-decoding test-time optimization (Ours w/o AD), and the best-performing baseline (TC+AE) introduced in \sect{exp:baselines}.
We find that when the target view is outside the training distribution and vastly different from the agent view, our full method shows a much better performance in achieving the target configuration. The variant without auto-decoding optimization and TC+AE fail to accomplish the task and exhibit an apparent deviation from the ground truth in the 3D points space of the fluids and floating cube.
We also provide quantitative comparisons on the control results. \fig{fig:quanti_control} shows the performance in the FluidPour and FluidShake environments.
We find our full model significantly outperforms the baseline approaches in both environments under all scenarios and evaluation metrics. The results effectively demonstrate the advantages of the learned 3D-aware scene representations, which contain a more precise encoding of the contents in the 3D environments and are capable of extrapolation viewpoint generalization.

\vspace{-2pt}
\subsection{Dynamic Prediction and Novel View Synthesis}
\vspace{-3pt}
Conditioned on a scene representation and an input action sequence, our dynamics model $f_\text{dyn}$ can iteratively predict the evolution of the scene representations.
Our decoder can then take the predicted state representation and reconstruct the corresponding visual observation from a query viewpoint. 
\fig{fig:quali_dynamics} shows that our model can accurately predict the future and perform novel view synthesis on four environments involving both fluid and rigid objects, suggesting its usefulness in trajectory optimization.
Please check our video results in the supplementary material for better visualization.
\vspace{-2pt}
\section{Conclusion}
\vspace{-3pt}
In this paper, we proposed to learn viewpoint-invariant 3D-aware scene representations from visual observations using an autoencoding framework augmented with a neural radiance field rendering module and time contrastive learning.
We show that the learned 3D representations perform well on the model-based visuomotor control tasks. When coupled with an auto-decoding test-time optimization mechanism, our method allows goal specification from a viewpoint outside the training distribution.
We demonstrate the applicability of the proposed framework in a range of complicated physics environments involving rigid objects and fluids, which we hope can facilitate future works on visuomotor control for complex and dynamic 3D manipulation tasks.


\clearpage


\bibliography{references,example}  

\newpage
\appendix

\vspace{-2pt}
\section{Model Details}
\label{apx:model_details}
\vspace{-3pt}

\begin{figure*}[h]
    \begin{center}
        \includegraphics[width=.6\linewidth]{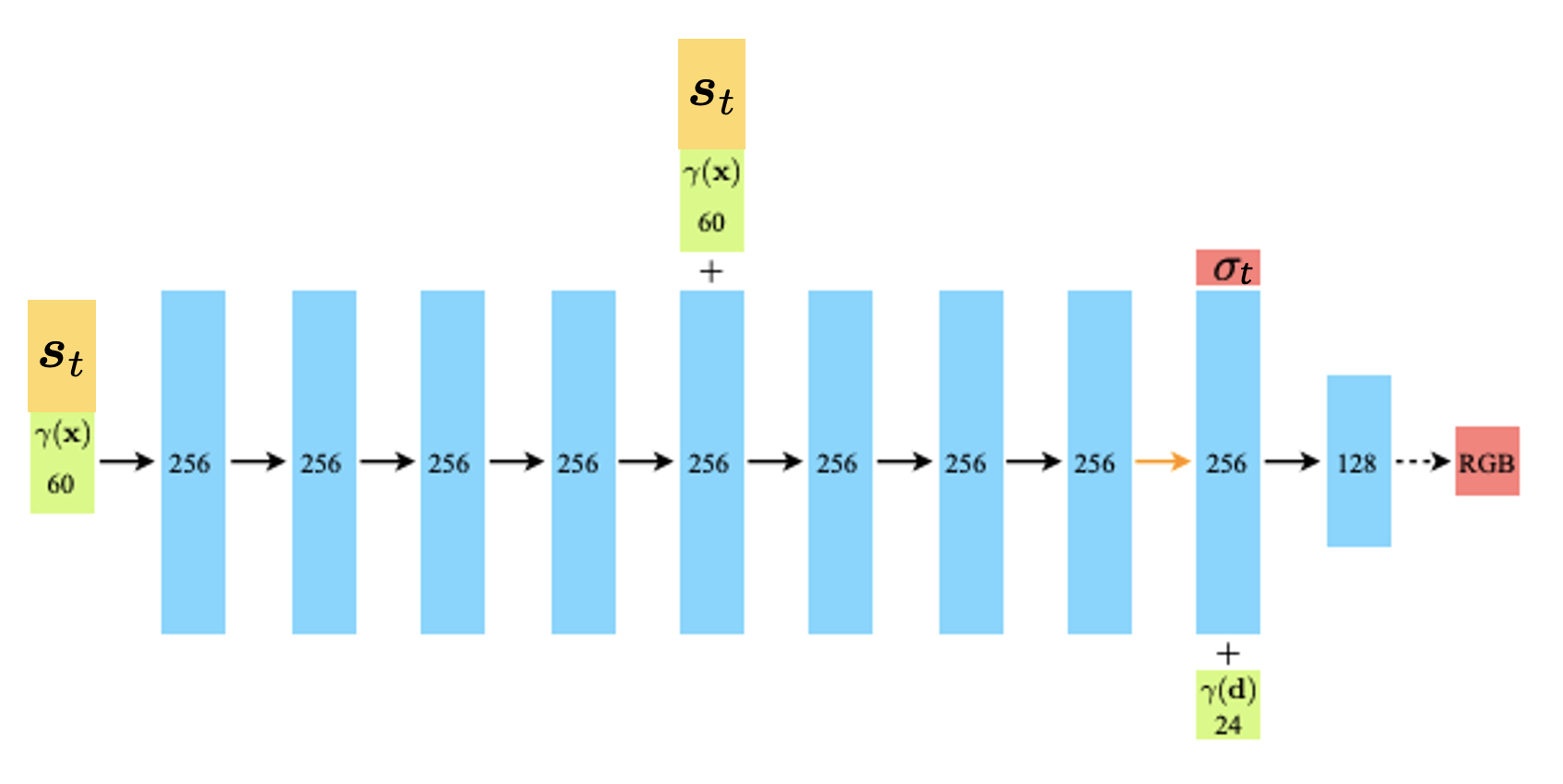}
    \end{center}
    \vspace{-8pt}
    \caption{\small{ {\bf A visualization of our decoder network architecture.}
    All layers are standard fully-connected layers, where the black arrows indicate layers with ReLU activations, orange arrows indicate layers with no activation, dashed black arrows indicate layers with sigmoid activation, and ``+" denotes vector concatenation.
    We concatenate the positional encoding of the input location $\gamma(\vx)$ and our learned state representation $\bm{s}_t$, and pass them through 8 fully-connected ReLU layers, each with 256 channels.
    We follow the NeRF~\cite{mildenhall2020nerf} architecture and include a skip connection that concatenates this input to the fifth layer’s activation.
    An additional layer outputs the volume density $\sigma_t$ (which is rectified using a ReLU to ensure that the output volume density is nonnegative) and a 256-dimensional feature vector.
    This feature vector is concatenated with the positional encoding of the input viewing direction $\gamma(\vd)$, and is processed by an additional fully-connected ReLU layer with 128 channels.
    A final layer (with a sigmoid activation) outputs the emitted RGB radiance at position $\vx$, as viewed by a ray with direction $\vd$, given the current 3D state representation $\bm{s}_t$.
    }
    }
    \vspace{-5pt}
    \label{apx_fig:decoder}
\end{figure*}

In the decoder model, we use a similar network architecture as the NeRF paper~\cite{mildenhall2020nerf}.
As shown in \fig{apx_fig:decoder}, we send a 3D point $\vx \in \mathbb{R}^3$, a camera ray $\vd \in \mathbb{R}^3$, and a state feature representation $\rvs_t$ into a fully-connected network and output the corresponding density $\sigma_t$ and RGB color $\vc_t$.

\vspace{-2pt}
\section{Environment Details}
\label{apx:environment_details}
\vspace{-3pt}

In the {\bf FluidPour} environment, we generated $1,000$ trajectories for training. Each trajectory has $300$ frames with $20$ camera views sampled around the objects with a fixed distance towards the world origin. The action space for the control task is the position and tilting angle of the cup, which are randomly generated when constructing the training set.

In the {\bf FluidShake} environment, we generated $1,000$ trajectories for training. Each trajectory has $300$ frames with $20$ camera views sampled around the objects with a fixed distance towards the world origin. The action space for the control task is the 2D location of the container in the world coordinate, which is also randomly generated when constructing the training set.

\begin{figure*}[t]
    \begin{center}
        \includegraphics[width=\linewidth]{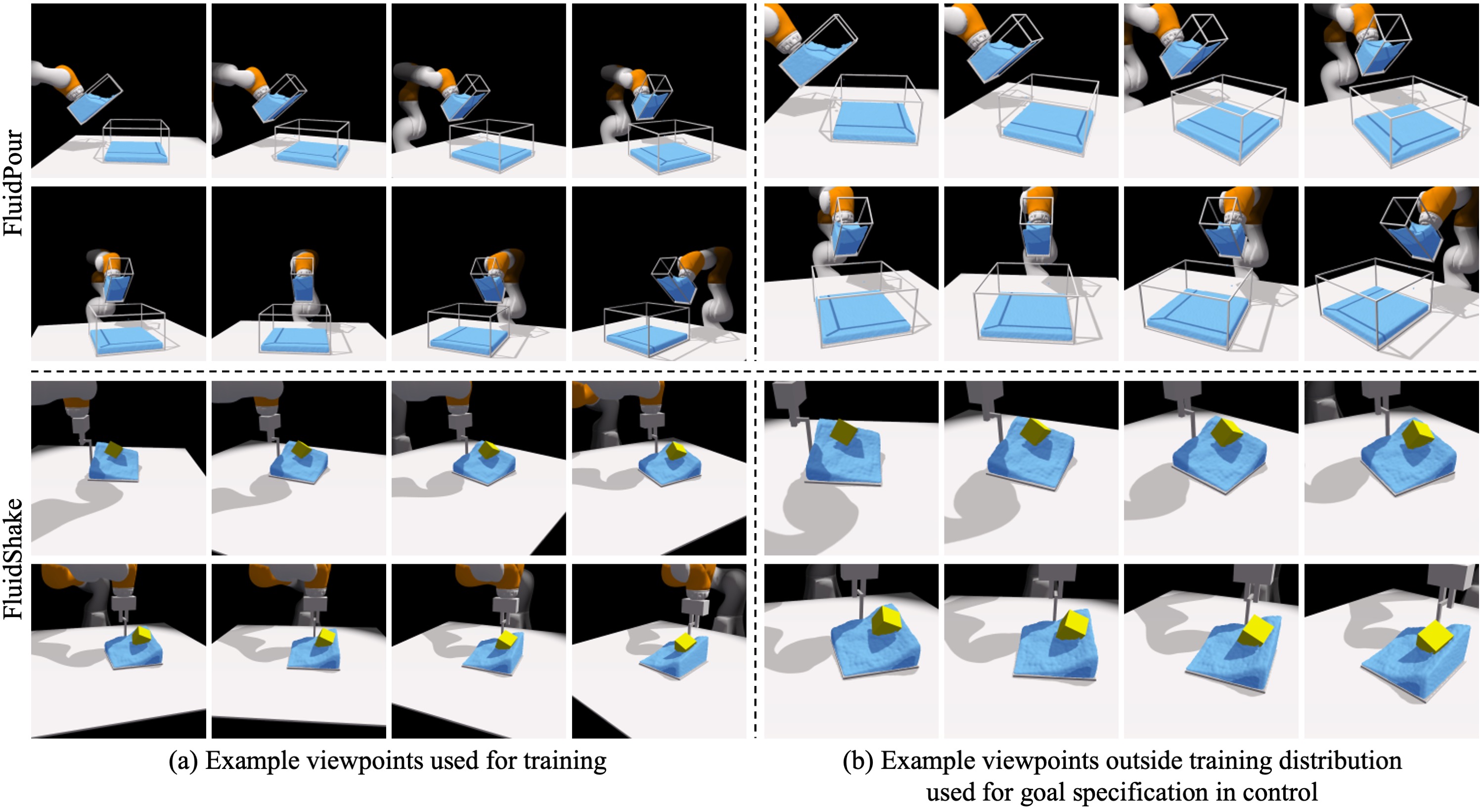}
    \end{center}
    \vspace{-8pt}
    \caption{\small { {\bf Comparison between the viewpoints used for training and the subsequent viewpoint extrapolation experiments.}
    (a) We show some example images that the model used during training. For both environments, the camera is placed from a fixed distance and facing towards the world origin.
    (b) To evaluate the model's ability for viewpoint extrapolation, \ie, processing visual observations from viewpoints that are outside the training distribution, we generate another set of viewpoints that are closer, higher, and facing more downwards.
    It is clear from the figure that the images from viewpoints used during training are very different from the ones used for viewpoint extrapolation when measured using pixel difference. Therefore, it is essential to build a model that can directly reason over 3D to provide the desired extrapolation generalization ability.
    Although the model has access to visual observations from multiple cameras during training, it can only observe the environment from one camera when performing the downstream control task.
    }
    }
    \vspace{-5pt}
    \label{apx_fig:viewpoint}
\end{figure*}

\fig{apx_fig:viewpoint} shows some example visual observations for FluidPour and FluidShake used during training. We also include some example images from viewpoints outside the training distribution, which are then used to evaluate our model's extrapolated generalization ability.

In {\bf RigidStack}, we generated $800$ trajectories for training. Each trajectory has $80$ frames with $20$ camera views sampled around the objects with a fixed distance towards the world origin.

In {\bf RigidDrop}, we generated $1,000$ trajectories for training. Each trajectory has $50$ frames with $20$ camera views sampled around the objects with a fixed distance towards the world origin.

\vspace{-2pt}
\section{Training Details}
\label{apx:training_details}
\vspace{-3pt}

For the encoder and decoder model described in \fig{fig:overview_train},
we use the Adam optimizer with the initial learning rate $5e^{-4}$ and decreased to $5e^{-5}$ for all the experiments. The batch size is $2$.
The hyperparameters in the decoder are the same as the original NeRF model \cite{mildenhall2020nerf} except the near and far distance between the objects and cameras are different in our environments.
In our FluidPour environment, we have $\text{near}=2.0$ and $\text{far}=9.5$.
In our FluidShake environment, we have $\text{near}=2.0$ and $\text{far}=7.0$.
In our RigidStack environment, we have $\text{near}=2.0$ and $\text{far}=7.0$.
In our FluidPour environment, we have $\text{near}=2.0$ and $\text{far}=6.0$.

\vspace{-2pt}
\section{Control Details}
\vspace{-3pt}

As discussed in \sect{sec:control}, we use model-predictive path integral (MPPI)~\cite{williams2015model} to solve the MPC problem.
MPPI is a sampling-based, gradient-free optimizer that considers temporal coordination between time steps when sampling action trajectories.
At time $t$, the algorithm first samples $M$ action sequences based on the current actions
$\va_t, \dots, \va_{T-1}$ via $ \hat{\va}^k_h = \va_h + \vn^k_{h}, k \in \{ 1, \dots, M\}, h \in \{ t, \dots, T - 1\} $.
Each noise sample $\vn^k_h$, denoting the noise value at the $h^\text{th}$ time step of the $k^\text{th}$ trajectory, is generated using filtering coefficient $\beta$ as the following:
\begin{equation}
\begin{aligned}
\small
    & \vu^k_{h} \sim \mathcal{N}(0, \Sigma) \quad \forall k \in \{ 1, \dots, M\}, h\in \{t, \dots, T-1 \} \\
    & \vn^k_{h} = \beta \cdot \vu^k_{h} + (1 - \beta) \cdot \vn^k_{h-1}, \text{where } \vn^k_{h<t} = 0.
\end{aligned}
\end{equation}
We then roll them out in parallel using the learned model on the GPU to derive $\hat{\bm{s}}_T^k, k \in \{ 1, \dots, M\}$, and then re-weight the trajectories according to the reward to update the action sequence using a reward-weighting factor $\gamma$:
$\va_h = (\sum_{k=1}^M\exp{(\gamma\cdot R^k)} \cdot \hat{\va}^k_h) / (\sum_{k=1}^M \exp{(\gamma \cdot R^k)}), h \in \{ t, \dots, T - 1\}$,
where $ R^k = - \| \hat{\bm{s}}^k_T - \bm{s}^\text{goal}  \|_2^2$.
This procedure is repeated for $L$ iterations at which point the best action sequence is selected.

The number of updating iteration for the auto-decoding test-time optimization $K$ is $500$. The number of sampled trajectories $M$ during MPPI optimization is set to $1,000$. The number of iterations $L$ for updating the action sequence is set to $100$ for the first time step, and $10$ for the subsequent control steps to maintain a better trade-off between efficiency and effectiveness. The reward-weighting factor $\gamma$ is set to $50$ and the filtering coefficient $\beta$ is specified as $0.7$. The control horizon $T$ is set as $80$ both for FluidPour and FluidShake. The hyperparameters are the same for all compared methods.

\vspace{-2pt}
\section{Additional Experimental Results}
\vspace{-3pt}

\subsection{Auto-Decoding Test-Time Optimization for Viewpoint Extrapolation}
\vspace{-3pt}

\begin{figure*}[t]
    \begin{center}
        \includegraphics[width=\linewidth]{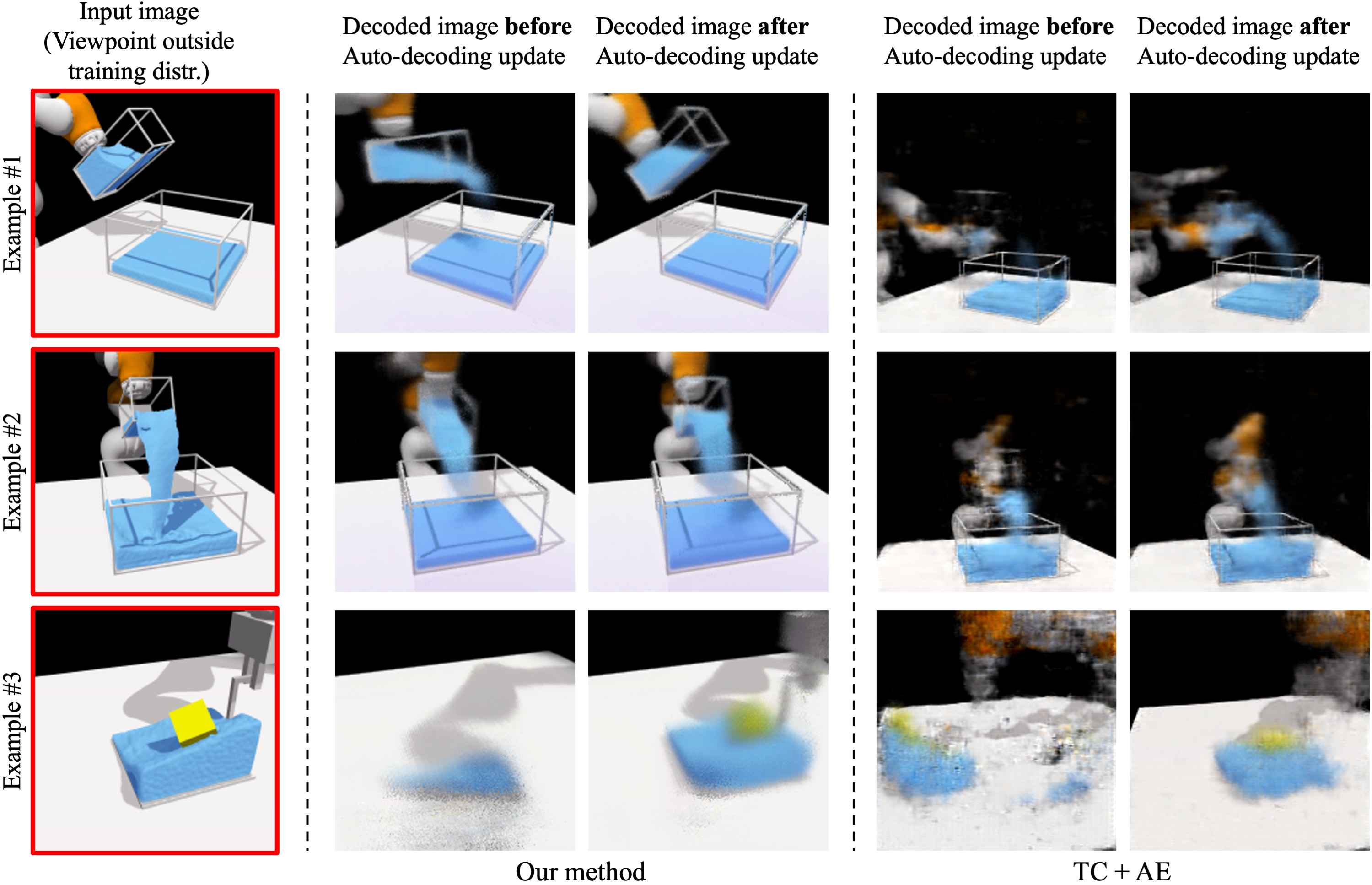}
    \end{center}
    \vspace{-8pt}
    \caption{\small { {\bf Qualitative results on auto-decoding test-time optimization.}
    Following from the pipeline illustrated in \fig{fig:overview_test}b,
    if the input image $I_t$ is outside the training distribution as shown on the left column, the encoder won't be able to generate the most accurate state representation.
    When passing the predicted state embedding $\bm{s}_t$ and the same viewpoint as the input to the decoder, the generated image does not match the underlying scene as shown in the second column.
    We then calculate the L2 distance of the pixels between the generated image and true observation, backpropagating the gradient until the state representation and making updates to $\bm{s}_t$ using SGD.
    As discussed in \sect{sec:autodecoder},
    the translational equivariance nature of the decoder allows it to effectively optimize the latent representation to make it better reflect the 3D contents in the scene.
    After the optimization, the generated visual observation is much closer to the ground truth, as shown in the third column.
    On the contrary, the vanilla autoencoder that uses a CNN-based decoder won't be able to capture the underlying scene even with test-time auto-decoding optimization, as shown on the right.
    }
    }
    \vspace{-5pt}
    \label{fig:auto_decoder}
\end{figure*}

\fig{fig:auto_decoder} shows the qualitative results on auto-decoding test-time optimization.
When encountering an image from a viewpoint outside the training distribution, this mechanism can help us derive a better representation of the scene that holds a more accurate description of the 3D contents.
The obtained representation after the optimization can then be used as the goal embedding $\bm{s}^\text{goal}$ in \eqn{eqn:planning}
that the agent needs to achieve.

\vspace{-2pt}
\subsection{Validation of the Dynamics Prediction on Real-World Data}
\vspace{-3pt}

\begin{figure}[t]
    \begin{center}
        \includegraphics[width=\linewidth]{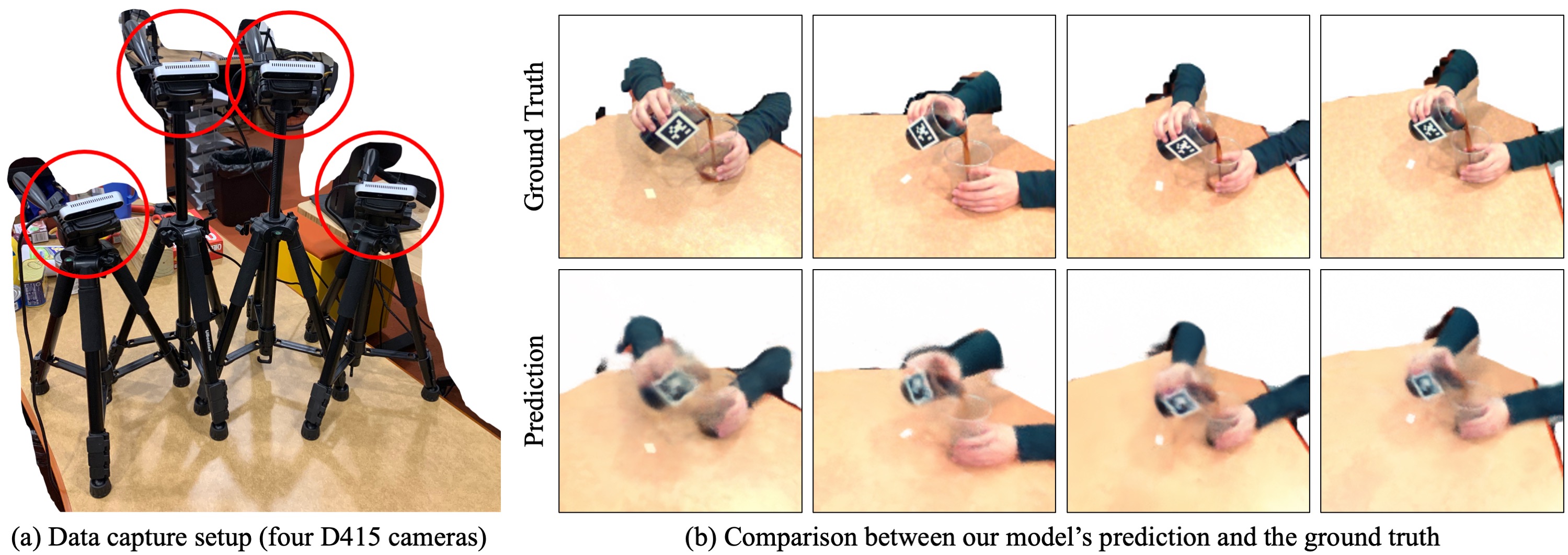}
    \end{center}
    \vspace{-8pt}
    \caption{\small{ {\bf Dynamics prediction using real-world data.}
    (a) We build a data recording set up containing four D415 RGBD cameras to record a human subject pouring water from one cup to another and then use the recorded data to evaluate our model's dynamics prediction ability.
    (b) We show a side-by-side comparison between the ground truth data and our model's prediction when predicting the future from the four camera views. Our model correctly identifies when the fluids pour out and start to fill the bottom container. Please see our supplementary video for better visualizations.
    }
    }
    \vspace{-0pt}
    \label{fig:real_world}
\end{figure}

We further evaluate our model's dynamics prediction ability by conducting experiments on real-world data. As shown in \fig{fig:real_world},
we use four D415 RGBD cameras to record a human subject pouring water from one cup to another. The cameras are calibrated and synchronized with the frequency of 15 Hz. We recorded 50 pouring episodes of length 15 seconds, resulting in a dataset of ~43,400 frames. We use the first 45 episodes for training and the remaining for testing.

To obtain the action, \ie, the movement of the cup that pours water out, we attached an AprilTag
~\cite{olson2011apriltag}
on the cup to obtain the 6 DoF pose of the cup at each time step.
From the supplementary video, you can see that our model can make open-loop future predictions on the testing trajectories in the representations space, i.e., given a latent scene representation of the current time step, the subsequent action sequence, our dynamic model can accurately predict the evolution of the latent scene representation and render the corresponding frames that closely resemble the ground truth.

\vspace{-2pt}
\subsection{Comparison With a PID Baseline That Only Matches the Robot’s State}
\vspace{-3pt}

To show the necessity of modeling the fluid dynamics in our task, we include an additional baseline that uses PID control~\cite{franklin2002feedback}
to reach the robot state in the target image without worrying about the fluid. Note that this baseline uses additional information, including the ground truth state of the robot in both the current and the target image.

Our supplementary video shows a controlling trial where the goal is to leave a small amount of fluid in the container at the end of the control episode.
Naively matching the container's position and orientation will not work, as shown in the PID baseline that the top container did not pour any fluid into the bottom container.
In contrast, our model first learns to pour a certain amount of fluid out and then tilt the container back to match the target configuration.
We further use the Chamfer distance to measure the models' performance in matching the fluid shape in the 3D point space, where our method significantly outperforms the PID baseline (Ours: 0.048237 vs. PID: 0.085727).

\begin{figure}[t]
    \begin{center}
        \includegraphics[width=\linewidth]{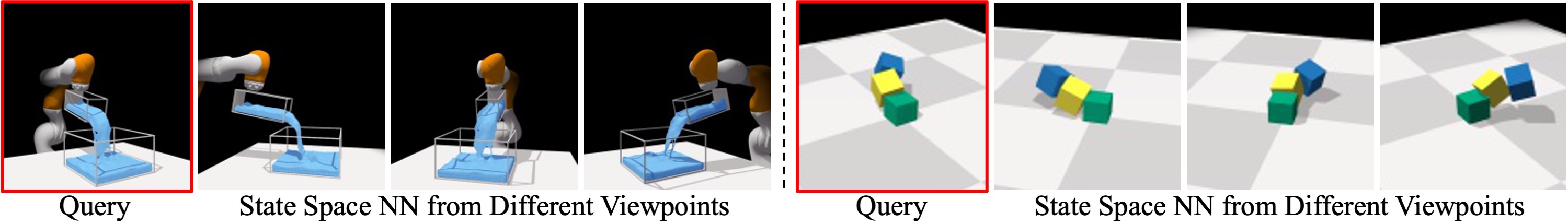}
    \end{center}
    \vspace{-8pt}
    \caption{\small{ {\bf Nearest neighbor (NN) results using our learned state representation.}
    Given a query image (red boundary), we search its nearest neighbors based on their state representation.
    Our learned scene representations can retrieve reasonable neighbor images, indicating that our state representations retain a good estimation of the contents inside the 3D scene and are invariant to camera poses.}
    }
    \vspace{-5pt}
    \label{fig:nearest_neighbor}
\end{figure}

\vspace{-2pt}
\subsection{Nearest Neighbor Search Using the Learned Representations}
\vspace{-3pt}

When the goal image is specified from a viewpoint different from the agent's view, to ensure the planning problem defined in \eqn{eqn:planning} still work, it is essential that the distance in the learned feature space reflects the distance in the actual 3D space, \ie, scenes that are more similar in the real 3D space should be closer in the learned feature space, even if the visual observations are captured from different viewpoints.
We visualize the nearest neighbor results in \fig{fig:nearest_neighbor}. Given a query image, we search its nearest neighbors based on their state representation $\bm{s}$ (introduced in \sect{sec:nerf_dynamic}).
Even if the images look quite different from each other when measured in pixel difference, the learned 3D-aware scene representations can retrieve reasonable neighborhood images that share similar 3D contents, indicating that the learned 3D-aware scene representations hold a good understanding of the real 3D scene and are invariant to viewpoint variations.

We conducted additional experiments to quantitatively measure the accuracy in finding the nearest neighbors (NN) across viewpoints of our proposed method. 
The results are shown in Table~\ref{tab:nearest_neighbor}.
Specifically, for each query frame, the model is asked to find the closest frame from a randomly selected viewpoint from a trajectory with $300$ frames. We measure the accuracy using the L1 distance between the time indexes. Randomly selecting the closest frame leads to an averaged distance of 100 times steps between the selected frame and the ground truth frame. Selecting based on the pixel difference is even worse. Our method can accurately find the nearest neighbor. The average time step difference between the selected frame and the ground truth frame is 1.772718 in FluidPour and 2.584928 in FluidShake.

\begin{table}
\centering
\begin{tabular}{ c | c | c | c }
 Env & Ours & NN in pixel space & Random \\ [0.5ex]
 \hline
 FluidPour & {\bf 1.772718} & 147.971993 & 99.903261 \rule{0pt}{2.6ex} \\ [0.5ex]
 FluidShake & {\bf 2.584928} & 132.467998 & 99.646617 \\
\end{tabular}
\vspace{5pt}
\caption{ \small {\bf Quantitative results on nearest neighbor (NN) search from different viewpoints.} We calculate the accuracy of finding NN using the L1 distance between the ground truth and retrieved time indexes. Our method measures the distance in the learned representation space, which delivers the best performance, and the retrieved frame is, on average, around two time steps away from the ground truth.
}
\label{tab:nearest_neighbor}
\vspace{-5pt}
\end{table}

\vspace{-2pt}
\section{Limitations and Future Works}
\vspace{-3pt}

Similar to many other data-driven methods, our model can deliver reasonable performance in regions well-supported by the data. However, for cases that our model has never seen before, \eg, more containers than during training, we wouldn't expect our model to generalize. Potential solutions may include (1) adding the examples and increasing the diversity of the training set or (2) using a more structured/compositional representation space instead of a single vector as in our model, which we leave for future works.

\end{document}